\documentclass[a4paper,fleqn]{cas-sc}

\usepackage[authoryear,longnamesfirst]{natbib}
\usepackage {enumitem}
\usepackage {enumerate}
\usepackage{graphicx}
\usepackage{float}
\usepackage{lineno}
\usepackage{perpage} 
\usepackage{natbib}
\MakePerPage{footnote}
\setcitestyle{numbers,square}

\def\tsc#1{\csdef{#1}{\textsc{\lowercase{#1}}\xspace}}
\tsc{WGM}
\tsc{QE}
\tsc{EP}
\tsc{PMS}
\tsc{BEC}
\tsc{DE}

\begin{document}
\let\WriteBookmarks\relax
\def\floatpagepagefraction{1}
\def\textpagefraction{.001}

\shorttitle{Fully Unsupervised Cross-lingual POS Tagging}

\shortauthors{Jianyu Zheng}

\title [mode = title]{Unsupervised Cross-Lingual Part-of-Speech Tagging with Monolingual Corpora Only}                      


%
\author[1,2]{Jianyu Zheng}[style=chinese,
                         orcid=0009-0007-1810-5010]

\cormark[1]


\ead{zhengjianyu@uestc.edu.cn}


\affiliation[1]{organization={School of Foreign Languages, University of Electronic Science and Technology of China},
    addressline={Chengdu}, 
    city={Sichuan Province},
    postcode={611731 CN}, 
    country={China}}
\affiliation[2]{organization={School of Computer Science and Engineering, University of Electronic Science and Technology of China},
    addressline={Chengdu}, 
    city={Sichuan Province},
    postcode={611731 CN}, 
    country={China}}

\cortext[cor1]{Corresponding author}

\begin{abstract}
Due to the scarcity of part-of-speech annotated data, existing studies on low-resource languages typically adopt unsupervised approaches for POS tagging. Among these, \textit{POS tag projection with word alignment} method transfers POS tags from a high-resource source language to a low-resource target language based on parallel corpora, making it particularly suitable for low-resource language settings. However, this approach relies heavily on parallel corpora, which are often unavailable for many low-resource languages. To overcome this limitation, we propose a fully unsupervised cross-lingual part-of-speech(POS) tagging framework that relies solely on monolingual corpora by leveraging unsupervised neural machine translation(UNMT) system. This UNMT system first translates sentences from a high-resource language into a low-resource one, thereby constructing pseudo-parallel sentence pairs. Then, we train a POS tagger for the target language following the standard projection procedure based on word alignments. Moreover, we propose a multi-source projection technique to calibrate the projected POS tags on the target side, enhancing to train a more effective POS tagger. We evaluate our framework on 28 language pairs, covering four source languages (English, German, Spanish and French) and seven target languages (Afrikaans, Basque, Finnis, Indonesian, Lithuanian, Portuguese and Turkish). Experimental results show that our method can achieve performance comparable to the baseline cross-lingual POS tagger with parallel sentence pairs, and even exceeds it for certain target languages. Furthermore, our proposed multi-source projection technique further boosts performance, yielding an average improvement of 1.3\% over previous methods.
\end{abstract}



\maketitle

\section{Introduction}
Processing for low-resource languages has been a major focus in the natural language processing(NLP) community\citep{pakray2025natural}. Among all NLP tasks, Part-of-Speech(POS) is one of the most crucial, as it forms the foundation for more advanced NLP applications.

\begin{figure*}[pos=ht]
\centering
\includegraphics[width=0.9\textwidth,height=0.2\textwidth]{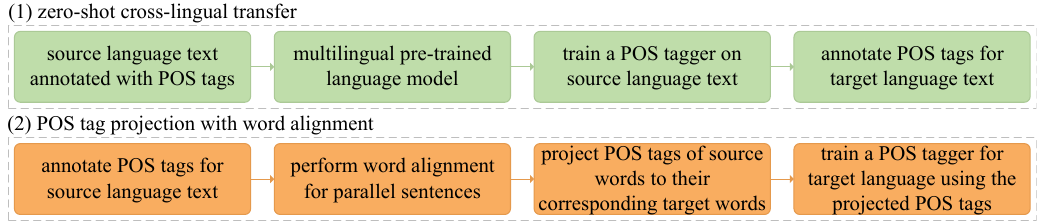} 
\caption{The steps of commonly used unsupervised cross-lingual POS tagging methods: (a) \textit{zero-shot cross-lingual transfer} using multilingual pre-trained language models; (b) \textit{POS tag projection through word alignment} with parallel corpus.}
\label{fig.1}
\end{figure*}

However, due to the lack of annotated POS tagging data in low-resource languages, numerous studies usually adopt unsupervised cross-lingual transfer techniques, which transfer POS knowledge from high-resource languages(such as English) to those low-resource languages, to facilitate POS tagging for these languages\citep{pires2019multilingual, eskander2020unsupervised}. As presented in Figure ~\ref{fig.1}, the current methods for unsupervised cross-lingual POS tagging can be categorized into two main types: (1) \textbf{zero-shot cross-lingual transfer} based on the multilingual pre-trained language models\citep{pires2019multilingual}; and (2) \textbf{POS tag projection with word alignment} using parallel corpus\citep{david2001inducing, fossum2005automatically, das2011unsupervised, duong2013simpler, agic2015if, tackstrom2013token, eskander2020unsupervised, eskander2022unsupervised}. Specifically, the former method uses POS annotation knowledge from source languages to fine-tune the language models, which are then directly applied to annotate the POS tags for the target language texts; While the latter aligns words across parallel texts, and then projects POS tags from the source-language words to their corresponding target-language words based on the alignment relationships. Subsequently, the projected instances on the target side are used to train a POS tagger for this language.

Compared to the "zero-shot cross-lingual transfer" method, "POS tag projection with word alignment" method holds many advantages. Since this method does not rely on pre-trained language models, it is more \textbf{lightweight} in both training and deployment, and \textbf{better generalization}, which means it can be applied to a wider range of low-resource languages. Moreover, it remains \textbf{relatively robust} when there exist significant morphological differences between the source and target languages\citep{eskander2020unsupervised}. However, the "POS tag projection with word alignment" method heavily relies on parallel corpora. For many low-resource languages, it is often difficult to obtain such corpora with high resource languages\citep{chopra2024zero}.

To address this issue, we propose a fully unsupervised cross-lingual part-of-speech(POS) tagging framework that relies solely on monolingual corpora. Specifically, we initially develop an unsupervised neural machine translation(UNMT) systems\citep{lample2018unsupervised, lample2018phrase, artetxe2017unsupervised, artetxe2019effective} by collecting large-scale monolingual data for both the source and target languages. Utilizing this UNMT system, source language sentences are translated into target language, constructing the pseudo-parallel sentence pairs. Then by adopting the above-mentioned “POS tag projection with word alignment” method, we can obtain the instances for training a POS tagger in the target language. Notably, to mitigate the issues caused by relying only on single source language during projection, such as sparse alignment and incorrect POS tag projections, we further introduce a multi-source projection strategy, which combines the annotation knowledge from several mutually parallel source languages to enhance the projection quality. Finally, based on the projected instances, we train a POS tagger for the target language using a BiLSTM architecture.

In the experiments, we evaluate a total of 28 language pairs, comprising 4 source languages (English, German, Spanish and French) and 7 target languages (Afrikaans, Basque, Finnis, Indonesian, Lithuanian, Portuguese and Turkish). By conducting experiments on the test sets of these target languages from the Universal Dependencies(UD)\citep{de2021universal}, our framework can achieve POS tagging accuracies exceeding 60\% for all target languages, comparable to the baseline method that relies on parallel corpora in the ”POS tag projection with word alignment” method\citep{eskander2020unsupervised, eskander2022unsupervised}. Moreover, for low-resource languages that are typologically similar to the source languages, such as Portuguese, Indonesian and Afrikaans, our framework can achieve accuracies of 92.0\%, 87.1\% and 89.5\%, respectively, outperforming the baseline with parallel corpora by 2.6\%$\sim$3.3\%. Additionally, the proposed multi-source projection technique further improves POS tagging accuracy, with gains of up to 0.6\%.

In this work, our contribution are as follows:
\begin{itemize}
\item [(1)] 
We propose a fully unsupervised cross-lingual POS tagging framework which relies solely on monolingual corpora. With the help of UNMT, we remove the reliance on parallel corpora required in the "POS tag projection with word alignment" method, our framework allows to implement cross-lingual POS tagging for any language pairs.
\item [(2)]
We introduce a revised calibration technique (multi-source projection strategy) for POS tag projection by considering multiple source languages. This technique can effectively alleviate sparse alignment and incorrect projection arising from single language pairs, thereby improving the quality of projected results for the target language, and enhancing the final POS tagging performance.
\item [(3)]
Through extensive experiments across a wide range of language pairs, we demonstrate the effectiveness of our proposed framework. The results show that our approach achieves performance comparable to the baseline "POS tag projection with word alignment" method requiring parallel corpora, and even outperforms the baseline in several target languages.
\end{itemize}

\section{Related work}
\subsection{Unsupervised cross-lingual POS Tagging}
The "POS tag projection with word alignment" method for unsupervised cross-lingual POS tagging was first proposed by Yarowsky et al.\citep{david2001inducing}. In their work, they employed denoising and smoothing techniques to mitigate null alignments in parallel sentence pairs. Based on the projection results, they trained a POS tagger for the target language using a Hidden Markov Model (HMM). Subsequent studies extended this approach by introducing techniques such as multi-source projection\citep{fossum2005automatically, agic2015if}, self-training\citep{duong2013simpler} and token and type constraints\citep{tackstrom2013token}, to address the issues related to null alignments, misalignments, and inconsistent POS tags resulting from language differences. Additionally, some studies focused on specific language pairs. For instance, Sukhareva et al.\citep{sukhareva2017distantly}  investigated the cross-lingual POS tagging for the German-Hittite pair, while Huck et al.\citep{huck2019cross}  focused on the Russian-Ukrainian pair.

Recently, Eskander et al.\citep{eskander2020unsupervised} successfully combined the multi-source projection and token and type constraints techniques to generate the training instances for POS tagging task in target languages. Then they adopted a revised BLSTM architecture, integrating pre-trained word representations, subword representations, and word cluster representation as input features, to train a POS tagger. Drawing from their work, our approach aims to perform unsupervised cross-lingual POS tagging using only monolingual corpora.

\subsection{Unsupervised machine translation}
Unsupervised machine translation (UMT) refers to training translation systems between two languages without relying on parallel corpora\citep{lample2018unsupervised}. In the deep learning era, most UMT approaches adopt neural network-based methods\citep{artetxe2017unsupervised, lample2018unsupervised}, commonly known as unsupervised neural machine translation(UNMT). UNMT first learns cross-lingual word representation in an unsupervised manner\citep{artetxe2017learning, artetxe2018robust, conneau2017word}, establishing a shared embedding space across languages. Using these shared representations, UNMT can learn translation mappings between language pairs through techniques such as denoising, autoencoding and back-translation\citep{sennrich2016improving}, etc.

To further enhance the performance of UNMT systems, many efforts have focused on optimizing the architecture of the MT systems. For example, Guillaume Lample et al.\citep{lample2018phrase} designed an encoder-decoder architecture that implements translation through techniques like denoising autoencoding and back translation. This approach not only significantly reduces the number of training parameters, but also enables the MT system to effectively learn translation relationships between languages. Other works attempt to improve the initialization the MT system parameters. For example, Lample and Conneau\citep{lample2019cross} utilized pre-trained language models to initialize the encoder and decoder modules in UNMT systems, improving the translation performance greatly.

Additionally, there are efforts to develop "one-to-many" NUMT systems. Sen et al.\citep{sen2019multilingual} designed an architecture featuring a language-agnostic encoder and language-specific decoders, enabling translation from English to multiple target languages. Sun et al. \citep{sun2020knowledge} developed a multilingual UNMT system using model distillation technique with an architecture that includes one encoder and one decoder. Furthermore, some work have explored leveraging auxiliary languages to improve multilingual UNMT. Garcia et al.\citep{garcia2021harnessing} utilized multiple high-resource languages with parallel corpora to assist in training the system. This approach enables the system to implicitly learn shared subword representation across languages, improving its generalization competence when translating between different language pairs. 

By comprehensively considering these MT system's performance and reproducibility, we select Guillaume Lample's work\citep{artetxe2019effective} to build our UNMT system.

\section{Method}
\begin{figure*}[pos=ht]
\centering
\includegraphics[width=0.8\textwidth,height=0.45\textwidth]{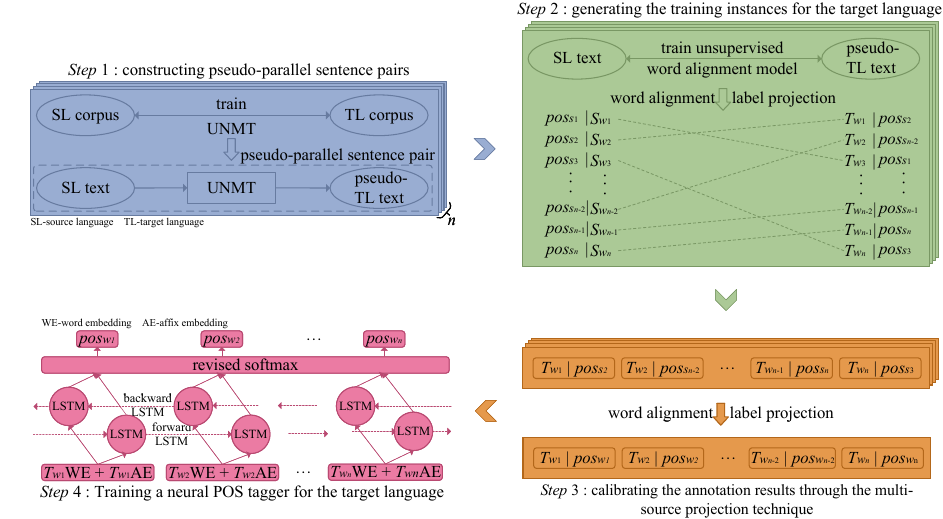} 
\caption{The fully unsupervised cross-lingual part-of-speech (POS) tagging framework, which does not rely on parallel sentence pairs. This framework consists four steps: 1) constructing pseudo-parallel sentence pairs; 2) generating the training instances for the target language; 3) calibrating the annotation results through the multi-source projection technique; 4) training a neural POS tagger for the target language.}
\label{fig.2}
\end{figure*}

The aim of our work is to explore how to train a fully unsupervised cross-lingual POS tagger using monolingual corpora only. As shown in Figure ~\ref{fig.2}, our framework consists of four steps: 1) constructing pseudo-parallel sentence pairs; 2) generating training instances with POS tag annotation for the target language; 3) calibrating the annotation results through the multi-source projection technique; and 4) training a neural POS tagger for the target language. In the following, we will describe each step in detail.

\subsection{Constructing pseudo-parallel sentence pairs}
For low-resource languages not supported by pre-trained multilingual language models, if they lack parallel sentence pairs with high-resource languages, the "POS tag projection with word alignment" method cannot be applied. To alleviate the reliance on the parallel sentence pairs, we propose to build a UNMT system trained solely on monolingual corpora from different languages, and then perform the unsupervised cross-lingual POS tagging task. Considering that it is relatively easy to obtain large-scale monolingual corpora from various languages, we use these corpora to train UNMT systems across various language pairs. Then utilizing these systems, we further translate sentences from high-resource languages into low-resource languages, thereby generating pseudo-parallel sentence pairs.

Considering both performance and reproduction, we adopt the method of training UNMT system proposed by Lample et al.\citep{lample2018phrase}, which consists of an encoder and a decoder. During the initialization phrase, they utilize the BPE algorithm\citep{sennrich2016neural} to process the corpora of both the source and target languages jointly, creating a shared subword vocabulary. This shared vocabulary enables words from both languages to be represented in a common space. After that, denoising autoencoding and back-translation techniques are applied iteratively to train the UNMT system. Specifically, during the denoising autoencoding phrase, a noise function $C$(·) is applied to introduce noise to both the source language text $x$ and the target language text $y$ . The encoder then maps these noised texts from different languages into a shared high-dimensional vector space. The decoder attempts to reconstruct their original representations from this shared representation; In the back translation phrase, the MT system ${\mathbf{M}}^{(t-1)}$ from the previous round translates the source sentence $x$ ( or the target sentence $y$) into an equivalent target sentence ${v}^{*}(x)$ (or source language sentence ${u}^{*}(y)$. The decoder attempts to reconstruct the source / target sentence as closely as possible. The objective function during training is to minimize the total loss from both denoising autoencoding and back-translation.

\subsection{Generating the training instances for the target language}
After generating a sufficient number of pseudo-parallel sentence pairs, we apply the "POS tag projection with word alignment" method to obtain annotation instances for training a POS tagger in the target language. During this step, we draw from the work of Eskander et al\citep{eskander2020unsupervised}, which achieves the state-of-the-art in this category. A brief introduction of their method is provided below:

After pre-processing the parallel sentence pairs, this work first adopt an off-the-shelf POS tagger to annotate the source language sentences. Then, utilizing an unsupervised word alignment tool, such as GIZA++, they perform bidirectional word alignment between source and target language sentences. By setting a threshold for the alignment probability, denoted as $\alpha$, they filter out low-quality aligned sentence pairs. Subsequently, following to token and type constraints\citep{tackstrom2013token}, they project the POS tags from the source words into the corresponding words in the aligned target language sentences. Finally, high-quality training instances are selected on the target side by considering factors such as the percentage of tokens with projected tags and the average alignment probability of word pairs within each sentence. These selected instances are used to train the POS tagger for the target language.

\subsection{Calibrating the annotation results through the multi-source projection technique}
According to the findings of Željko Agić et al\citep{agic2016multilingual}, monolingual alignment mapping is prone to several issues, such as null alignment and incorrect tag mapping, etc. As a result, many studies\citep{fossum2005automatically, agic2015if, eskander2020unsupervised, eskander2022unsupervised} have utilized multiple source languages that are parallel to each other to improve the projection results of POS tags on the target side. We also adopt this technique. However, unlike previous work, which use genuine parallel sentence pairs from various source languages to align sentences within the same target language, our generated target language sentences differ in forms. This is because the UNMT systems are trained on difference language pairs, each system with various levels of performance. Therefore, directly following the methods from previous work could be infeasible in our case.

However, training a MT system still needs to adhere to the "equivalence principle"\citep{stahlberg2020neural} when transforming text between languages. This principle reflects that the generated target language text should not only conform to the grammar rules of the target language, but also maintain equivalence with the source language in terms of lexicon, syntax and semantics, etc\citep{nida2021toward}. Based on this principle , we find that there is a significant amount of overlapping words between these generated sentences to ensure the "equivalence principle" is satisfied. As illustrated in Table ~\ref{table.1}, we display four pseudo-parallel sentence pairs with the source sentences in English, German, French and Spanish, respectively. These sentences are translated into Portuguese using the UNMT systems we have trained. Although the expressions in the translated sentences differ to some extent due to variations in system performance, we observe that many words overlap, such as "metodologia", "mesma", and "que". Therefore, we decide to calibrate the POS tags for these "overlapping words" in order to improve the projection quality.

When calibrating the annotated results from projection, we assume that once a word in the target language sentence has been assigned a specific POS tag, these tags reflect a high degree of confidence. Hence, we adopt maximum voting according to the results to determine the POS tags of these "overlapping words" \footnote{We previously attempted to identify the projection results based on alignment confidence, but it did not yield a significant difference.}. The calculation formula is displayed as follows:
$$tag(w)={argmax}_{tag}\textstyle\sum_{i,s}^{}p({tag}_{i,s}|w),$$
where $w$ represents the overlapping words for the target side, and $p({tag}_{i,s})$ takes values \{0 | 1\}, indicating whether a word projected from source language $s$, can be tagged with POS tag $i$. Subsequent experiments reveal that we should select the "psudo" target sentences performing best under monolingual mapping, to carry out the multi-source projection technique and calibrate the annotation results on such sentences, so that the POS tagger trained on these sentences yields better performance.

\begin{table}[]
\caption{Four pseudo-parallel sentence pairs. The source language sentences are mutually parallel, which are in English, German, French and Spanish, respectively. After these source sentences are translated into Portuguese using different UNMT system, we highlight the "overlapping words" by underlining them in the target sentences.}
\label{table.1}
\begin{tabular}{cll}
\hline
\textbf{\begin{tabular}[c]{@{}c@{}}Source \\ language\end{tabular}} & \multicolumn{1}{c}{\textbf{Source sentences}}& \multicolumn{1}{c}{\textbf{Translated sentences in Portuguese}}\\
\hline
English & \begin{tabular}[c]{@{}l@{}}We were told that there had to be a study using \\ exactly the same methodology .\end{tabular}& \begin{tabular}[c]{@{}l@{}}Fomos {\underline{informados}} \underline{de} \underline{que} havia \underline{que} ter \underline{um} \underline{estudo}\\ usando \underline{exatamente} \underline{a} \underline{mesma} \underline{metodologia}.\end{tabular} \\
German & \begin{tabular}[c]{@{}l@{}}Uns wurde mitgeteilt , es müsse eine Studie geben ,\\ die exakt die gleiche Methodik verwendet .\end{tabular} & \begin{tabular}[c]{@{}l@{}}Uns foram \underline{informados} , deveriam haver \underline{uma} pesquisa\\ \underline{que} \underline{precisas} \underline{de} usar \underline{a} \underline{mesma} \underline{metodologia} .\end{tabular} \\
French& \begin{tabular}[c]{@{}l@{}}Ensuite, qu' il fallait une étude qui utilise\\ exactement la même méthodologie .\end{tabular}& \begin{tabular}[c]{@{}l@{}}Assim , \underline{que} só \underline{uma} \underline{pesquisa} \underline{que} utiliza \\ \underline{exatamente} \underline{a} \underline{mesma} \underline{metodologia} .\end{tabular}\\
Spanish & \begin{tabular}[c]{@{}l@{}}Se nos dijo que debía haber un estudio que \\ utilizara exactamente la misma metodología .\end{tabular}& \begin{tabular}[c]{@{}l@{}}Ele nos disse \underline{que} devia haver \underline{um} \underline{estudo} \underline{que}\\ utilizasse \underline{exatamente} \underline{a} \underline{mesma} \underline{metodologia}.\end{tabular}\\
\hline                 
\end{tabular}
\end{table}

\subsection{Training a neural POS tagger for the target language}
After obtaining high-quality training annotation instances for training a POS tagger on the target side, we adopt a revised BLSTM+softmax\citep{graves2012long} architecture to train the POS tagger. The BLSTM is widely used in sequence labeling tasks due to its effectiveness and simplicity, while the softmax function is used to decode the output of BLSTM. Notably, since some words in the training instances are tagged as NULL, we set the output value of the corresponding neurons to -$\infty$, ensuring that the loss of these positions is not calculated during model training. Unlike the work of Eskander et al\citep{eskander2020unsupervised}, our model input consists solely of randomly initialized word representations and their corresponding affix n-gram character representations, without considering information, like brown word clusters\citep{brown1992class} or contextual words' representations, as we find that the inclusion of these information led to a decrease in performance in our experiments. We hypothesize that this may be due to the fact that the target sentences are pseudo sentences generated by UNMT system, which differ from natural sentences. Adding extra information could negatively impact the model's performance.

\section{Experiment setup}
\subsection{Dataset}
We evaluate our framework on a total of 28 language pairs, comprising 4 common source languages and 7 target languages. We represent each language using its ISO 639-1 code. A detailed description of these languages is provided in Table~\ref{table.2}.

For building UNMT systems, we randomly choose 10 million sentences for each language from the CC-100 corpus\footnote{https://huggingface.co/datasets/cc100}. Additionally, 3000 parallel sentence pairs from the CCMatrix\footnote{https://github.com/facebookresearch/LASER/tree/master/tasks/CCMatrix} and GNOME dataset\footnote{https://cswww.essex.ac.uk/Research/nle/corpora/GNOME/} are randomly chosen as the validation and test sets\footnote{The validation and test sets for the five language pairs—"German-Basque", "Spanish-Afrikaans", "French-Afrikaans", "French-Basque" and "French-Finnish"—are chosen from the GNOME dataset, while that of the remaining language pairs are sourced from the CCMatrix dataset}. Notably, we have verified that there is no overlap between the sentences in the training, validation and test sets for each language pair.

To construct pseudo-parallel sentence pairs, we use the Europarl corpus\footnote{https://www.statmt.org/europarl/} as the source language text and translate it into the corresponding target language. The Europarl Parallel Corpus is extracted from the proceedings of the European Parliament, which covers the four source languages used in our experiment, with sentences being parallel to one another. After preprocessing, over 1.5 million sentences pairs are obtained for each source language. We then randomly choose 100 thousand sentences from this corpus for translation.

For evaluating the POS taggers we have trained, we use the test set from Universal Dependencies\citep{de2021universal} for the target languages\footnote{Afrikaans-AfriBooms, Basque-BDT, Finnish-TDT, Indonesian-GSD, Lithuanian-ALKSNIS, Portuguese-Bosque and Turkish-IMST}. 

\begin{table}[]
\caption{The introduction of languages used in our experiment, which includes four high-resource languages and seven low-resource languages. The language family and group for each language is also displayed in this table.}
\label{table.2}
\begin{tabular}{cll}
\hline
\multicolumn{2}{c}{\textbf{Languages}}& \multicolumn{1}{c}{\textbf{Language Group}}  \\
\hline
\multirow{4}{*}{\begin{tabular}[c]{@{}c@{}}Source \\ languages\end{tabular}} & English & Indo-European family, Germanic group \\
& Spanish& Indo-European family, Romance group\\
& French & Indo-European family, Romance group\\
& Germn& Indo-European family, Germanic group\\
\hline
\multirow{7}{*}{\begin{tabular}[c]{@{}c@{}}Target\\ languages\end{tabular}}  & Afrikaans  & Indo-European family, Germanic group\\
& Basque& Isolated language\\
& Finnish& Uralic family, Finno-Ugric group\\
& Indonesian & Austronesian family, Malayo-polynesian group\\
& Lithuanian & Indo-European family, Baltic group\\
& Portuguese & Indo-European family, Romance group\\
& Turkish& Altaic family, Turkic group\\
\hline
\end{tabular}
\end{table}

For the selection of experimental tools, similar to Eskander et al's work\citep{eskander2020unsupervised}, we use GIZA++\footnote{https://www2.statmt.org/moses/giza/GIZA++.html} for word alignment, and Stanza\footnote{https://github.com/stanfordnlp/stanza} for POS tagging the source language sentences. Notably, Stanza supports the four source languages in our experiment and follows the Universal Dependency annotation guidelines to annotate the POS tags of the source language texts.

\subsection{Implementation details}
The parameter settings for training the UNMT systems(as described in Section 3.1) are displayed in Table ~\ref{table.3}. For key parameters in the steps about training POS taggers, we follow the settings from Eskander et al 's work\citep{eskander2020unsupervised}. For example, during implementing the bidirectional word alignment in section 3.2, we set the alignment confidence $\alpha$ to 0.1 . Furthermore, the parameter settings for the "BLSTM+softmax" architecture( as described section 3.4) are presented in Table~\ref{table.4}.   
 
\begin{table}[]
\caption{Parameter settings for training UNMT systems.}
\label{table.3}
\begin{tabular}{ccccc}
\hline
\textbf{parameters}   & \textbf{values} & \textbf{} & \textbf{parameters}& \textbf{values} \\
\hline
optimizer & adam &  & learning rate& 1e-4\\
encoder layers& 4& & decoder layers& 4 \\
shared encoder layers & 3 & & shared decoder layers & 3 \\
word shuffle& 3& & word dropout & 0.1\\
word blank& 0.2 & & back-translation coefficient & 1 \\
batch size& 32 & & epoch size & 100  \\
early stopping  & 8 &  &\\
\hline
\end{tabular}
\end{table}

\begin{table}[]
\caption{Parameter settings for training POS taggers.}
\label{table.4}
\begin{tabular}{ccccc}
\hline
\textbf{parameters} & \textbf{values} & \textbf{} & \textbf{parameters}& \textbf{values} \\
\hline
optimizer & adam & & regularization & L2 \\
learning rate & 1e-3  & & learning decay rate& 0.1 \\
word embedding size & 64 &  & affix embedding size & 64 \\
epoch size & 20 & & number of nodes in hidden layer & 128 \\
dropout rate & 0.7 & & number of dropout layers  & 2 \\
\hline
\end{tabular}
\end{table}

\subsection{Baselines}
We select the following two baselines for comparison with our framework. Notably, our focus is on developing a fully unsupervised cross-lingual POS tagger using only monolingual corpora. Therefore, we aim to assess the extent to which our method can achieve performance of the baselines.

\noindent{\textbf{Cross-lingual POS tagger with parallel sentence pairs}   The unsupervised cross-lingual POS tagging method proposed by Eskander et al\citep{eskander2020unsupervised} forms the basis of our approach. Specifically, they use parallel sentence pairs from the Bible corpus\footnote{http://christos-c.com/bible/} for word alignment and POS tag projection. However, the Bible corpus, being a religious text, is out of domain. It consists of over 31,000 sentences, with the New Testament alone accounting for more than 7,900 sentences. The domain and scale of this corpus may limit the performance of the POS tagger they trained to some extent.}

\noindent{\textbf{Supervised POS tagger}  We select Stanza\citep{qi2020stanza} for comparison. Stanza is one of the most representative supervised POS taggers, trained on Universal Dependencies annotation dataset\citep{de2021universal}, and can be used as an upper bound for comparison with our method.}

\subsection{Evaluation metrics}
We use BLEU\citep{papineni2002bleu} to evaluate the performance of UNMT systems, and the macro-F1 score to assess the performance of the POS tagger for each target language.

\section{Results and analysis}
\subsection{Main experimental results}
Table~\ref{table.5} displays the performance of UNMT systems on the testing set for 28 language pairs. Since these UNMT systems are only used to generate pseudo-parallel sentence pairs, we do not compare their performance with that of other similar MT systems. From the results in Table ~\ref{table.5}, we observe significant performance differences across language pairs, with language relationship as a crucial influencing factor. For example, the UNMT systems achieve higher BLEU scores for pairs like "English/German→Afrikaans"(Indo-European family, Germanic group) and "Spanish/French→Portuguese"(Indo-European family, Romance group). In addition, the UNMT system performs well when translating from English to Indonesian, likely because Indonesian has incorporated numerous English loanwords due to western culture influence. English and Indonesian also share a similar alphabet, differing primarily in pronunciation\citep{andi2013comparative}. In contrast, languages like Basque (Isolated language) and Turkish (Turkic group), which are more distantly related to the source languages, yielding poor performance.

\begin{table}[]
\caption{The performance of UNMT systems with BLEU scores for each language pair, including the translation orientations from the source to the target languages and vice versa. The best performance for each language in the translation process is highlighted in bold.}
\label{table.5}
\begin{tabular}{clccccc}
\hline
\multicolumn{1}{l}{}&& \multicolumn{1}{l}{\textbf{English}} & \multicolumn{1}{l}{\textbf{Spanish}} & \multicolumn{1}{l}{\textbf{French}} & \multicolumn{1}{l}{\textbf{German}} & \multicolumn{1}{l}{\textbf{Ave}} \\
\hline
\multirow{2}{*}{Afrikaans}& src→tgt &\textbf{41.5}&17.7&21.0&32.6&28.2\\
& tgt→src&44.0&9.6&11.6&\textbf{25.2}&22.6\\
\hline
\multirow{2}{*}{Basque}& src→tgt&5.6& \textbf{12.7}&4.6&6.8& 7.4\\
& tgt→src&8.2&15.6& 4.7& 4.8& 8.3\\
\hline
\multirow{2}{*}{Finnish}& src→tgt&\textbf{10.5}&7.8&6.1&9.4& 8.5\\
&tgt→src&17.4&9.7&3.4&9.8&10.1\\
\hline
\multicolumn{1}{l}{\multirow{2}{*}{Indonesian}} & src→tgt & \textbf{34.3}& 23.3&17.2 & 13.5& 22.1\\
\multicolumn{1}{l}{}& tgt→src & 36.0& 20.7& 14.6& 10.6& 20.5\\
\hline
\multicolumn{1}{l}{\multirow{2}{*}{Lithuanian}} & src→tgt & \textbf{11.1}& 10.1&9.5& 10.4&10.3\\
\multicolumn{1}{l}{}& tgt→src&18.0& 14.5& 12.6&11.1&14.1\\
\hline
\multicolumn{1}{l}{\multirow{2}{*}{Portuguese}} & src→tgt&41.5& \textbf{61.3}& 37.8& 21.8& 40.6\\
\multicolumn{1}{l}{}& tgt→src & \textbf{47.3}& \textbf{61.5}& \textbf{38.5}& 18.7& \textbf{41.5}\\
\hline
\multirow{2}{*}{Turkish}& src→tgt&5.8&\textbf{7.1}&5.0& 6.2&6.0\\
& tgt→src&9.0&7.5&5.4&5.5&6.9\\
\hline
\multirow{2}{*}{Ave}& src→tgt&21.5&20.0&14.5&14.4& 17.6 \\
& tgt→src&25.7&19.9&13.0 & 12.2&17.7\\
\hline
\end{tabular}
\end{table}

Table~\ref{table.6} presents the POS tagging results for the seven target languages. Our method achieves tagging accuracy over 60\% across all languages. This indicates that a fully unsupervised POS tagging system, trained using only monolingual corpora, can achieve a reasonable level of accuracy that meets basic tagging requirements. Furthermore, we observe that language relationships also influence POS tagging accuracy. The POS tagger trained on the language pairs such as Afrikaans → English/German(Indo-European family, Germanic group) and Portuguese →Spanish/French(Indo-European family, Romance group) typically achieves strong performance. However, for the target languages like Basque(Isolated language) and Turkish(Altaic family, Turkic group), which differ significantly from the source languages in terms of language family, the POS taggers perform less effectively.

\begin{table}[]
\caption{POS tagging accuracy on the testing set of Universal Dependency(\%). In columns 2 to column 6, the first value in each cell represents the performance of our framework, while the value in the bracket below is the result of the cross-lingual POS tagger with parallel sentence pairs baseline\citep{eskander2020unsupervised}. The "multi-source" column shows the results using the multi-source projection technique, while the "Stanza" column indicates the performance of the supervised POS tagger baseline.}
\label{table.6}
\begin{tabular}{ccccccc}
\hline
\multicolumn{1}{l}{} & \textbf{English}& \textbf{Spanish}&\textbf{French}&\textbf{German}& \textbf{"multi-source"}& \textbf{Stanza} \\
\hline
Afrikaans& \begin{tabular}[c]{@{}c@{}}87.4\\ (86.9)\end{tabular}& \begin{tabular}[c]{@{}c@{}}80.8\\ (83.1)\end{tabular} & \begin{tabular}[c]{@{}c@{}}86.7\\ (83.9)\end{tabular}& \begin{tabular}[c]{@{}c@{}}\textbf{89.5}\\ (84.1)\end{tabular} & \begin{tabular}[c]{@{}c@{}}\textbf{89.5}\\ (\textbf{89.3})\end{tabular} & 97.6 \\
\hline
Basque& \begin{tabular}[c]{@{}c@{}}72.2\\ (\textbf{67.3})\end{tabular}& \begin{tabular}[c]{@{}c@{}}81.3\\ (64.6)\end{tabular} & \begin{tabular}[c]{@{}c@{}}71.4\\ (65.8)\end{tabular} & \begin{tabular}[c]{@{}c@{}}72.5\\ (66.7)\end{tabular}& \begin{tabular}[c]{@{}c@{}}\textbf{81.4}\\ (67.1)\end{tabular} & 96.2\\
\hline
Finnish& \begin{tabular}[c]{@{}c@{}}\textbf{80.7}\\ (82.8)\end{tabular} & \begin{tabular}[c]{@{}c@{}}73.7\\ (80.9)\end{tabular} &\begin{tabular}[c]{@{}c@{}}74.9\\ (80.0)\end{tabular} & \begin{tabular}[c]{@{}c@{}}72.0\\ (82.0)\end{tabular} &\begin{tabular}[c]{@{}c@{}}\textbf{80.7}\\ (\textbf{83.4})\end{tabular} & 97.0\\
\hline
Indonesian& \begin{tabular}[c]{@{}c@{}}87.1\\ (\textbf{84.1})\end{tabular}& \begin{tabular}[c]{@{}c@{}}83.4\\ (83.5)\end{tabular} & \begin{tabular}[c]{@{}c@{}}82.9\\ (82.9)\end{tabular} & \begin{tabular}[c]{@{}c@{}}79.9\\ (81.2)\end{tabular}& \begin{tabular}[c]{@{}c@{}}\textbf{87.3}\\ (83.0)\end{tabular} & 93.7\\
\hline
Lithuanian& \begin{tabular}[c]{@{}c@{}}77.8\\ (80.9)\end{tabular}& \begin{tabular}[c]{@{}c@{}}75.3\\ (78.2)\end{tabular} & \begin{tabular}[c]{@{}c@{}}73.8\\ (79.0)\end{tabular} & \begin{tabular}[c]{@{}c@{}}73.1\\ (78.7)\end{tabular} & \textbf{\begin{tabular}[c]{@{}c@{}}78.2\\ (82.5)\end{tabular}} & 93.4 \\
\hline
Portuguese& \begin{tabular}[c]{@{}c@{}}88.3\\ (86.1)\end{tabular}& \begin{tabular}[c]{@{}c@{}}92.0\\ (\textbf{88.7})\end{tabular} & \begin{tabular}[c]{@{}c@{}}91.5\\ (86.6)\end{tabular} & \begin{tabular}[c]{@{}c@{}}88.0\\ (81.2)\end{tabular}& \begin{tabular}[c]{@{}c@{}}\textbf{92.6}\\ (87.8)\end{tabular} & 97.0\\
\hline
Turkish& \begin{tabular}[c]{@{}c@{}}65.2\\ (74.3)\end{tabular}& \begin{tabular}[c]{@{}c@{}}66.0\\ (72.7)\end{tabular} & \begin{tabular}[c]{@{}c@{}}60.0\\ (\textbf{74.7})\end{tabular} & \begin{tabular}[c]{@{}c@{}}67.2\\ (72.8)\end{tabular}& \begin{tabular}[c]{@{}c@{}}\textbf{67.4}\\ (74.6)\end{tabular} & 94.2\\
\hline
Ave & \begin{tabular}[c]{@{}c@{}}79.8\\ (80.3)\end{tabular}& \begin{tabular}[c]{@{}c@{}}78.9\\ (78.8)\end{tabular} & \begin{tabular}[c]{@{}c@{}}77.3\\ (79.0)\end{tabular} & \begin{tabular}[c]{@{}c@{}}77.5\\ (78.1)\end{tabular}& \textbf{\begin{tabular}[c]{@{}c@{}}82.4\\ (81.1)\end{tabular}} & 95.6\\
\hline
\end{tabular}
\end{table}

Notably, compared to the other three source languages, English is easier to use for training a higher-performing POS tagger, achieving an average of POS tagging accuracy of 79.8\% in our experiment. This could be attributed to English's statue as a global lingual franca, with numerous annotation standards based on it, which are subsequently applied to other languages. Additionally, as global communication becomes more frequent, texts in other languages often incorporate English vocabulary (a phenomenon known as "code-switching" phenomenon), which benefits the training of UNMT systems, and thereby contributes to better-performing POS tagger.

The multi-source projection technique we propose can further enhance POS tagging accuracy. For example, this technique improves tagging accuracy by 0.6\% in Portuguese. Moreover, our technique outperforms Eskander et al‘s multi-source projection technique\citep{eskander2020unsupervised}, achieving an average improvement of 1.3\%. This suggests that combing multiple source languages is beneficial for improving the quality of training instances in the target language. 

Compared to other baselines, our method exhibits a performance gap with the supervised POS tagger baseline, Stanza. Besides, although our method does not perform as well as the cross-lingual POS tagger with parallel sentence pairs baseline\citep{eskander2020unsupervised}, it still achieves comparable results. Notably, for 12 out of the 28 language pairs, our method surpasses the baseline. This is because the target languages where our method exceeds the baseline are often closely related to the source languages, such as Afrikaans and Portuguese. The tagging accuracy of our method in these two target languages reaches 89.5\% and 92.0\%, respectively, outperforming the baselines by approximately 2.6\%$\sim$3.3\%. Additionally, our method also exceeds the baseline method in Basque. This is due to the fact the baseline uses a relatively small number of parallel sentences, consisting of only 7,900 sentences from the New Testament, when training the POS tagger for Basque. This highlights that Eskander et al’s method\citep{eskander2020unsupervised} could be more susceptible to the scale and genre of parallel sentences, whereas our method is more robust.

\subsection{Analysis}
\subsubsection{The UNMT’s performance \& the POS tagger's accuracy}
To investigate whether the performance of UNMT systems impacts the accuracy of the POS tagger for the target language, we conduct a relevance analysis across the 28 language pairs in our experiment. The results show a correlation coefficient of 0.8339, determined through a Pearson correlation test with $p$ < 0.05. This finding indicates a significant positive correlation between the performance of UNMT systems and the accuracy of POS taggers for the target languages.

\subsubsection{The effect of pseudo-parallel sentence pairs}
\begin{figure*}[pos=ht]
\centering
\includegraphics[width=0.52\textwidth,height=0.36\textwidth]{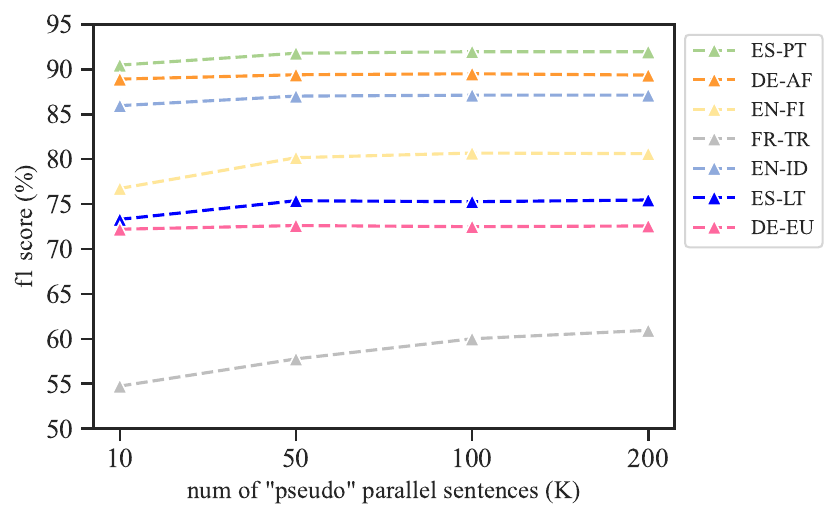} 
\caption{The effect of the number of pseudo-parallel sentence pairs on the accuracy of POS tagger for target languages. Seven language pairs are chosen for this experiment.}
\label{fig.3}
\end{figure*}

We also investigate the effect of the number of pseudo-parallel sentence pairs on the POS tagger accuracy for target languages. Seven language pairs are selected for the experiment, including "Spanish-Portuguese", "German-Afrikaans", "English-Finnish", "French-Turkish", "English-Indonesian", "Spanish-Lithuanian" and "German-Basque". For each language pair, we experiment with 10k, 50k, 100k, and 200k sentence pairs. As shown in Figure ~\ref{fig.3}, we find that as the number of pseudo-parallel sentence pairs increases from 10k to 50k, the accuracy of the POS taggers for the target languages improves to varying degrees. However, when the number of sentence pairs exceeds 50k, the improvements in accuracy become marginal(except for Turkish). These results indicate that training a POS tagger to achieve optimal performance requires a certain scale of pseudo-parallel sentence pairs. However, beyond a certain threshold, further increases in the number of sentence pairs does not result in significant improvements for the POS tagger's performance.

\subsubsection{Per POS-tag accuracy}
\begin{figure*}[pos=ht]
\centering
\includegraphics[width=0.90\textwidth,height=0.54\textwidth]{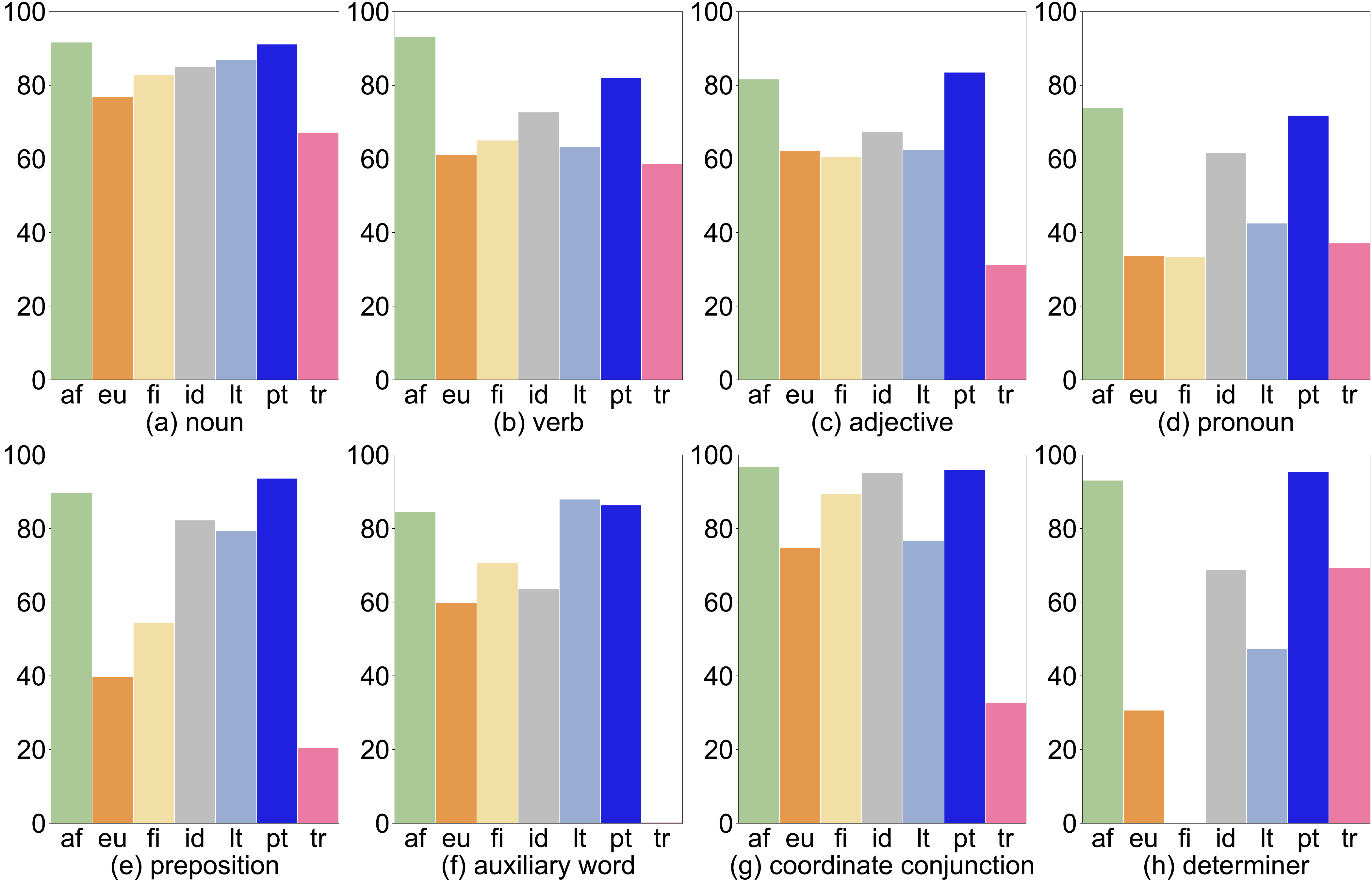} 
\caption{The accuracy of POS taggers for the seven target languages across each POS category. For each target language, the average accuracy of POS taggers trained on the four source languages is reported. The first row displays the accuracy for four content word categories (noun, verb, adjective and pronoun); while the second row shows the accuracy for four function word categories (preposition, auxiliary word, coordinating conjunction, and determiner). }
\label{fig.4}
\end{figure*}

We analyze the accuracy of the POS taggers for each POS category, covering four types of content words(noun, verb, adjective and pronouns) and four types of function words(prepositions, auxiliary words, coordinating conjunctions and determiners). As shown in Figure ~\ref{fig.4}, for each target language, we report the average performance of the POS taggers trained on the four source languages. The detailed performance of each POS tagger for each language pair is provided in Appendix ~\ref{Appendix.A}.

The accuracy of the content words follows the pattern: noun > verb $\approx$ adjective > pronoun. This ranking, however, may vary across specific languages. For example, in Turkish, pronouns have higher accuracy than adjectives. Additionally, by examining the training instances on the target side, we find that pronouns, which have the lowest accuracy, are more susceptible to misalignment and incorrect tag projection, leading to poorer prediction performance. Across all target languages, Afrikaans and Portuguese achieve higher accuracy for content words, while Turkish shows the lowest performance.

Among the four types of function words, coordinating conjunctions exhibit the highest accuracy, followed by prepositions and auxiliary words, while determiners show the lowest accuracy. Notably, for certain target languages, some categories of function words are predicted extremely poorly. For example, the accuracy for auxiliary words in Turkish is only 0.3. By observing the training instances in this language, we find that auxiliary words are highly prone to misalignment and incorrect tag projection. Furthermore, across all target languages, Afrikaans and Portuguese again show higher accuracy, whereas Turkish has the lowest performance.

By comparing the results between content words and function words, we find that, except for coordinate conjunctions, the overall prediction accuracy for function words is lower than that for content words. This discrepancy may be because, when using GIZA++ for word alignment, content words across languages are more align accurately, whereas function words often exhibit weaker or less consistent alignments.

Furthermore, we also analyze the performance of the POS taggers on multi-category words. Multi-category words refer to words that exhibit two or more POS characteristics, and can serve different grammatical roles depending on the contexts\citep{gupta2011tengram}. Strong performance on such words indicates that a POS tagger can effectively capture contextual information and accurately predict the appropriate POS tag for each word.

\begin{figure*}[pos=ht]
\centering
\includegraphics[width=0.85\textwidth,height=0.33\textwidth]{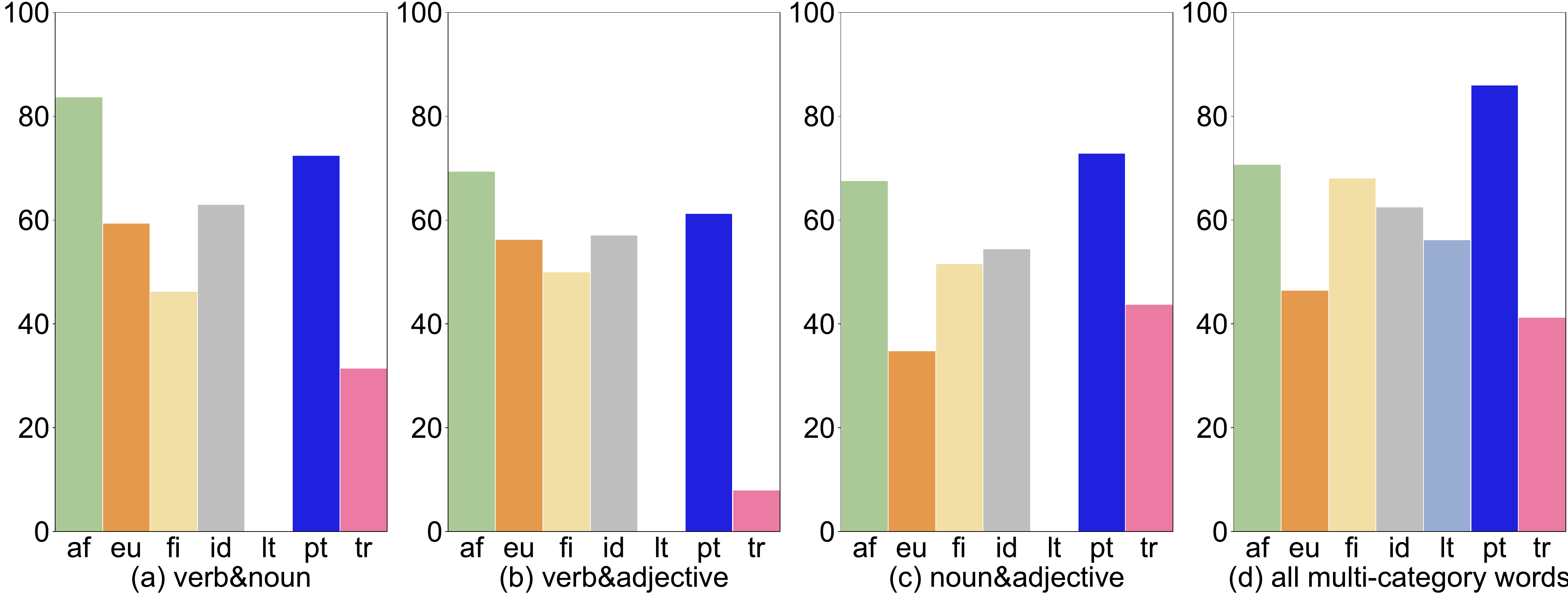} 
\caption{Accuracy of POS taggers for seven target languages on multi-category words. (a), (b) and (c) display the accuracy results for "verb\&noun", "verb\&adjective" and "noun\&adjective", respectively, while (d) shows the accuracy for all multi-category words.}
\label{fig.5}
\end{figure*}

The prediction accuracy of multi-category words for each target language is shown in Figure~\ref{fig.5}. Detailed prediction results for multi-category words in each language pair are provided in Appendix~\ref{Appendix.B}. As shown in Figure~\ref{fig.5}(d), Portuguese achieves the highest accuracy for multi-category words among all target languages, while Turkish demonstrates the lowest. Furthermore, Figures~\ref{fig.5}(a), (b), and (c) illustrate three representative types of multi-category words. Words categorized as "verb \& noun" achieve relatively high prediction accuracy, while "verb \& adjective" and "noun \& adjective" show varying performance depending on the target language. Notably, these three types of multi-category words do not appear in Lithuanian. This is because Lithuanian, one of the oldest languages in the modern Indo-European family, has relatively rigid word classes. Particularly, nouns and adjectives are less likely to serve other grammatical roles in different contexts.

\subsubsection{The effect of multi-source projection technique}
\begin{figure*}[pos=ht]
\centering
\includegraphics[width=0.55\textwidth,height=0.32\textwidth]{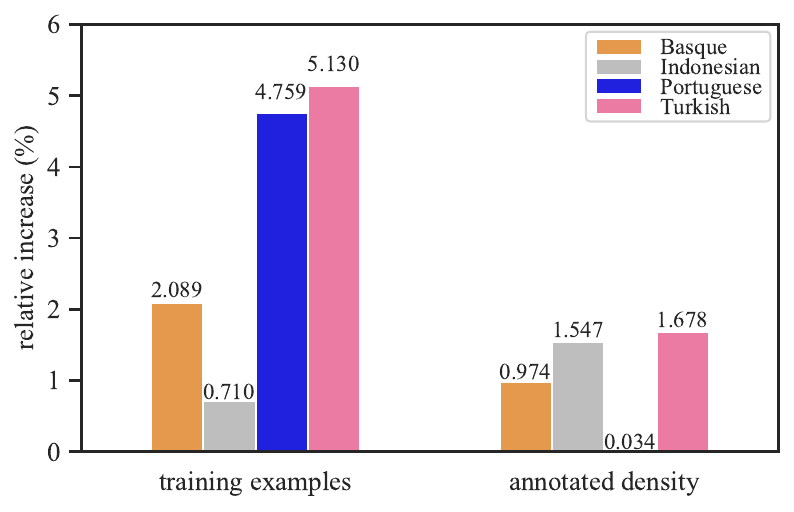} 
\caption{Relative increase in training examples and annotation density achieved through the multi-source projection technique for four target languages: Basque, Indonesian, Portuguese and Turkish.}
\label{fig.6}
\end{figure*}

According to Table~\ref{table.6}, our proposed multi-source projection technique improves POS tagging accuracy in 5 out of 7 target languages. By analyzing these five target languages, we find that for four of them(except Lithuanian), both the number of training instances and projected annotation densities increase through this technique, as illustrated in Figure~\ref{fig.6}. This enhancement contributes to training more effective POS tagger. For Lithuanian, although the number of training instances does not significantly increase, we observe that the POS tagging quality of words on the target side becomes more accurate after applying this technique. Consequently, POS tags and the co-occurrence patterns between words can be learned more effectively during training. As shown in the example sentence in Table~\ref{table.7}, the word \textit{kas} in the testing set is an out-of-vocabulary(OOV) word. However, the POS tagger trained on the text processed by the multi-source projection technique can better capture the contextual information, thereby correctly predicting the POS tag of this word.

\begin{table}[]
\caption{POS tag annotation results for an example Lithuanian sentence produced by different UNMT systems.}
\label{table.7}
\begin{tabular}{cl}
\hline
\textbf{UMT system}& \multicolumn{1}{c}{\textbf{Example sentence}}\\
\hline
EN—LT& Labai\_AUX čia\_ADV \textcolor{red}{kas\_DET} jo\_PRON prašė\_VERB kištis\_NOUN !\_PUNCT  \\
"Multi-source projection"& Labai\_PRON čia\_ADV \textcolor{green}{kas\_PRON} jo\_PRON prašė\_VERB kištis\_NOUN !\_PUNCT \\
Golden Set& Labai\_ADV čia\_PART kas\_PRON jo\_PRON prašė\_VERB kištis\_VERB !\_PUNCT\\
\hline
\end{tabular}
\end{table}

\subsubsection{Bad cases}
By examining the training instances and POS tag annotation results on the testing sets of the target languages, we summarize some typical error cases as follows:
\begin{itemize}
\item [(1)] 
Ineffective alignment of particle words. When using GIZA++ for word alignment, many particle words (such as articles and prepositions) are not accurately aligned. This problem is particularly evident in one-to-many and many-to-many alignment involving these words. For example, in the pseudo-parallel sentence pairs of English-Lithuanian, the Lithuanian particle "kad" is tagged \textit{NULL} in approximately 55\% of training cases. Consequently, particle words on the target side fail to receive POS tags through word alignment and annotation, causing the trained POS tagger to inaccurately predict their POS categories. 
\item [(2)]
Incorrect projection of POS tags. Due to cross-lingual differences in grammatical usage, even semantically equivalent words across languages may serve different POS roles within their respective sentences\citep{nikolaev2020fine}. For instance, in the English-Turkish language pair, the word \textit{kim} in Turkish often acts as a pronoun. However, through alignment and projection in our method, this word is frequently annotated as a particle, and is never labeled as a pronoun in the training instances.
\end{itemize}

Therefore, we will consider incorporating linguistic knowledge to address the aforementioned issue and further enhance the POS taggers’ performance.

\section{Conclusion}
In this work, we propose a fully unsupervised cross-lingual POS tagging framework with only monolingual corpora, leveraging unsupervised machine translation to alleviate the reliance on parallel corpora in the previous POS tag projection with word alignment methods. In experiments covering 28 language pairs, our framework can achieve performance comparable to the baseline method based on parallel sentence pairs. Notably, for target languages that are closely related to the source languages, our approach outperforms previous work, with improvements of 2.6\%$\sim$3.3\% in tagging accuracy. Furthermore, the multi-source projection technique we propose can further enhance the performance, achieving an average gain of 1.3\%.

In future work, we plan to employ model distillation to transfer richer linguistic knowledge from source language into the target language. Moreover, we utilize more advanced UNMT system to further enhance the quality of pseudo-parallel sentence pairs, thereby improving the overall performance of our POS tagging framework.

\vspace{1em}
\noindent{\textbf{Data Availability Statement}: All data that support the findings of this study are available from the corresponding author upon reasonable request.}

\vspace{1em}
\noindent{\textbf{Acknowledgement}: I thank Ying Liu at Tsinghua University for her helpful comments. I also thank the Center of High performance computing, Tsinghua University for providing computing resources.}

\vspace{1em}
\noindent{\textbf{Funding Statement}: This work was supported by Fundamental Research Funds for the Central Universities of Ministry of Education of China(Grant No. ZYGX2025WXJ007), the Postdoctoral Fellowship Program of CPSF (Grant No. GZC20252558), and the China Postdoctoral Science Foundation (Grant No. 2025M783842).}

\printcredits

\bibliographystyle{cas-model2-names}

\bibliography{output}

\appendix
\section{\textbf{Appendix A} \normalfont The accuracy of POS tags across 28 language pairs}
\centering
\label{Appendix.A}
\begin{tabular}{ccccccccc}
\hline
\textbf{} & \textbf{Noun} & \textbf{Verb} & \textbf{Adjective} & \textbf{Pronoun} & \textbf{Preposition} & \textbf{\begin{tabular}[c]{@{}c@{}}Auxiliary\\ Word\end{tabular}} & \textbf{\begin{tabular}[c]{@{}c@{}}Coordinate\\ Conjunction\end{tabular}} & \textbf{Determiner} \\
\hline
de-af&96.1&92.2&82.3&83.0&91.5&99.4&99.2& 95.2\\
en-af&91.4&95.4&88.1&93.3&80.4&96.6&98.2&93.9\\
es-af&88.2&93.1&76.1&31.2&93.5&74.0&92.9&89.6\\
fr-af&91.2&92.1&80.2&88.3&93.7&68.1&96.8&93.7\\
af\_ave&91.7&93.2&81.7&74.0&89.8&84.5&96.8&93.1\\
\hline
de-eu&75.7&48.9&53.4&20.2&26.2&66.5&74.9&39.6\\
en-eu&75.2&57.7&64.6&49.4&48.6&56.9&75.4&15.1\\
es-eu&77.8&73.2&70.8&40.5&51.6&69.9&74.2&54.4\\
fr-eu&78.4&64.7&60.1&25.0&32.6&46.7&75.0&13.8\\
eu\_ave&76.8&61.1&62.2&33.8&39.8&60.0&74.8&30.7\\
\hline
de-fi&83.4&45.1&67.0&22.7&49.6&63.9&92.3&-\\
en-fi&84.8&73.6&71.2&52.7&65.6&86.4&88.8&-\\
es-fi&79.1&70.5&53.3&23.4&56.5&72.0&90.3&-\\
fr-fi&84.1&71.2&51.3&34.9&46.7&60.9&86.2&-\\
fi\_ave&82.9&65.1&60.7&33.4&54.6&70.8&89.4&-\\
\hline
de-id&87.0&61.7&62.5&21.4&81.1&30.1&97.5&70.1\\
en-id&86.1&76.6&73.8&95.5&88.2&92.6&95.6&60.4\\
es-id&81.5&74.6&65.6&64.1&82.2&81.7&94.2&83.9\\
fr-id&85.6&77.8&67.2&65.2&77.6&50.7&93.1&61.3\\
id\_ave&85.1&72.7&67.3&61.6&82.3&63.8&95.1&68.9\\
\hline
de-lt&87.2&49.8&65.0&37.1&72.3&91.2&77.0&53.0\\
en-lt&87.1&68.0&69.5&65.0&85.9&90.2&76.6&35.5\\
es-lt&86.1&66.7&54.7&33.9&82.0&80.4&84.0&62.0\\
fr-lt&87.2&68.5&60.8&34.4&77.4&90.2&69.8&39.0\\
lt\_ave&86.9&63.3&62.5&42.6&79.4&88.0&76.8&47.4\\
\hline
de-pt&90.2&72.9&79.7&68.5&89.9&79.9&93.5&96.4\\
en-pt&89.4&84.2&80.7&68.2&89.2&92.8&98.8&93.4\\
es-pt&91.9&80.6&89.7&76.2&97.2&87.9&98.0&97.2\\
fr-pt&93.4&90.7&84.0&74.3&98.4&84.9&94.4&95.2\\
pt\_ave&91.2&82.1&83.5&71.8&93.7&86.4&96.1&95.5\\
\hline
de-tr&72.5&58.8&33.5&30.3&25.8&0.0&32.1&76.2\\
en-tr&68.6&55.0&37.4&46.1&19.2&0.5&29.0&66.3\\
es-tr& 68.5&63.5&28.9&36.0&15.8&0.5&37.5&64.8\\
fr-tr&59.1&57.3&25.1&36.5&21.4&0.0&32.4&70.7\\
tr\_ave&67.2&58.7&31.2&37.2&20.6&0.3&32.8&69.5\\
\hline
ave&83.1&70.9&64.2&50.6&65.7&64.8&80.3&67.5 \\
\hline
\end{tabular}

\section{\textbf{Appendix B} \normalfont The accuracy of POS tags across 28 language pairs}
\label{Appendix.B}
\begin{tabular}{ccccc}
\hline
\textbf{} & \multicolumn{1}{l}{\textbf{verb\&noun}} & \multicolumn{1}{l}{\textbf{verb\&adjective}} & \multicolumn{1}{l}{\textbf{noun\&adjective}} & \multicolumn{1}{l}{\textbf{All}} \\
\hline
de-af& 79.9&66.7&82.4&87.1\\
en-af&85.2&66.7&70.6&59.8\\
es-af&89.9&77.8&76.5&65.8\\
fr-af&79.9&66.7&41.2&69.9\\
af\_ave&83.7&69.4&67.6&70.7\\
\hline
de-eu&50.3&15.0&31.8&51.1\\
en-eu&51.7&80.0&34.6&42.9 \\
es-eu&72.8&90.0&41.7&53.4\\
fr-eu&62.9&40.0&31.3&38.4\\
eu\_ave&59.4&56.3&34.8&46.5\\
\hline
de-fi&34.1&50.0&53.1&64.2\\
en-fi&53.7&50.0&56.3&76.3\\
es-fi&46.3&50.0&40.6&68.6\\
fr-fi&51.2&50.0&56.3&63.2\\
fi\_ave& 46.3&50.0&51.6&68.1\\
\hline
de-id&59.3&59.5&50.7&48.0\\
en-id&61.1&57.1&58.9&73.4\\
es-id&63.0&57.1&49.3&64.8\\
fr-id& 68.5&54.8&58.9&63.7\\
id\_ave&63.0&57.1&54.5&62.5\\
\hline
de-lt&-&-&-&59.8\\
en-lt&-&-&-&57.0\\
es-lt&-&-&-&55.8\\
fr-lt&-&-&-&52.4\\
lt\_ave&-&-&-&56.2\\
\hline
de-pt&70.6&60.0&65.9&83.9\\
en-pt&69.7&55.0&63.1&84.0\\
es-pt&74.8&67.5&82.9&88.3\\
fr-pt&74.8&62.5&79.8&87.7\\
pt\_ave&72.5&61.3&72.9&86.0\\
\hline
de-tr&47.8&4.3&37.5&45.8\\
en-tr&43.5&8.5&50.0&42.1\\
es-tr&21.7&6.4&50.0&36.4\\
fr-tr&13.0&12.8&37.5&40.9\\
tr\_ave&31.5&8.0&43.8&41.3\\
\hline
ave&59.4&50.3&54.2&61.6\\
\hline
\end{tabular}

\end{document}